\documentclass[conference]{IEEEtran}
\IEEEoverridecommandlockouts

\usepackage{cite}
\usepackage{amsmath,amssymb,amsfonts}
\usepackage{algorithmic}
\usepackage{graphicx}
\usepackage{textcomp}
\usepackage{xcolor}
\usepackage[T1]{fontenc}
\usepackage{newtxtext,newtxmath}
\def\BibTeX{{\rm B\kern-.05em{\sc i\kern-.025em b}\kern-.08em
    T\kern-.1667em\lower.7ex\hbox{E}\kern-.125emX}}

\begin{document}

\title{Scalable multilingual PII annotation for responsible AI in LLMs}

\author{
\IEEEauthorblockN{
Bharti Meena,
Joanna Skubisz,
Harshit Rajgarhia, 
Nand Dave,
Kiran Ganesh, \\
Shivali Dalmia,
Abhishek Mukherji,
Vasudevan Sundarababu
}

\IEEEauthorblockA{
Centific, United States \\
\{bharti.meena, joanna.skubisz, harshit.rajgarhia, nandvinaykumar.dave, kiran.ganesh, \\
shivali.dalmia, abhishek.mukherji, vasudevan.sundarababu\}@centific.com}

}

\maketitle

\begin{abstract}
As Large Language Models (LLMs) gain wider adoption, ensuring their reliable handling of Personally Identifiable Information (PII) across diverse regulatory contexts has become essential. This work introduces a scalable multilingual data curation framework designed for high-quality PII annotation across 13 underrepresented locales (Table \ref{tab:localevolumes}), covering approximately 336 locale-specific PII types. Our phased, human-in-the-loop annotation methodology combines linguistic expertise with rigorous quality assurance, leading to substantial improvements in recall and false positive rates from pilot, training, and production phases. By leveraging inter-annotator agreement metrics and root-cause analysis, the framework systematically uncovers and resolves annotation inconsistencies, resulting in high-fidelity datasets suitable for supervised LLM fine-tuning. Beyond reporting empirical gains, we highlight common annotator challenges in multilingual PII labeling and demonstrate how iterative, analytics-driven pipelines can enhance both annotation quality and downstream model reliability.
\end{abstract}


\section{Introduction}
\label{sec:introduction}
\subsection{PII Data Protection}
\label{sec:introduction.motivation}

The surge in user-generated content has led to vast textual corpora containing hidden instances of Personally Identifiable Information (PII) in application forms, support tickets, reviews and social media posts \cite{baazarvoice}. 
PII—such as \texttt{NAME}, \texttt{SSN}, and \texttt{PHONE NUMBER}—poses significant privacy risks if not handled correctly. Its compromise can result in identity theft, financial fraud, and unauthorized access to sensitive data \cite{bigid}. 

To address these threats, regulations such as GDPR, HIPAA, and CCPA mandate strict safeguards and penalties for non-compliance \cite{gdpreu, szczepanska2025phi, oag}. However, existing automated PII annotation systems—often based on Named Entity Recognition (NER)—struggle with ambiguity, format variability, and low-resource languages. This necessitates high-quality multilingual annotations to fine-tune LLMs for robust PII labeling and redaction.

\subsection{Challenges in Data Annotation for LLM SFT}
\label{sec:introduction.challenges}

In the space of data annotation\footnote{Terms annotation and labeling are used interchangeably.} for Supervised Fine-Tuning (SFT) of LLMs, below is the set of challenges that we address in this work.

\begin{table}[b]
\centering
\resizebox{\columnwidth}{!}{%
\begin{tabular}{|l|c|c|c|}
\hline
\textbf{Language-Locale} & \textbf{Pilot} & \textbf{Training} & \textbf{Production} \\
 & \textbf{Volumes} & \textbf{Volumes} & \textbf{Volumes}\\ 

\hline
ar-UAE (Arabic - UAE) & 75-80 & 1000-1200 & 4000-4200 \\\hline
fi-FI (Finnish - Finland) & 95-100 & 1000-1200 & 4000-4200 \\\hline
hi-IN (Hindi - India) & 70-75 & 1000-1200 & 4000-4200 \\\hline
no-NO (Norwegian - Norway) & 150-170 & 1000-1200 & 4000-4200 \\\hline
nl-BE (Dutch - Belgium) & 25-30 & 300-350 & 1000-1200 \\\hline
nl-NL (Dutch - Netherlands) & 70-75 & 700-800 & 2500-2600 \\\hline
pl-PL (Polish - Poland) & 80-85 & 1000-1200 & 4000-4200 \\\hline
pt-BR (Portuguese - Brazil) & 50-55 & 700-800 & 2500-2600 \\\hline
pt-PT (Portuguese - Portugal) & 25-30 & 300-400 & 1000-1200 \\\hline
sv-SE (Swedish - Sweden) & 80-85 & 1000-1200 & 4000-4200 \\\hline
zh-CN (Chinese - China) & 75-80 & 700-800 & 2500-2600\\\hline
zh-SG (Chinese - Singapore) & 25-30 & 300-350 & 1000-1200\\
\hline
\end{tabular}%
}
\vspace{2mm}
\caption{Number of annotated tasks per \textit{Locale} in each phase.}
\label{tab:localevolumes}
\end{table}

\begin{enumerate}
\item While current LLMs continue to become more effective in English and a few handful of global languages, LLMs still lack basic understanding of several tier-2 underrepresented languages such as those listed in Table  \ref{tab:localevolumes}. Thus, expansion to these languages requires dedicated efforts in data annotation.
\item The study demonstrates that current large language models exhibit substantial limitations in detecting locale-specific PII, achieving a recall of no more than 65\% in the case of Chinese \cite{zeng2025automated}. Common issues include missed detections and incorrect PII type classifications, largely due to limited digital representations of such entities. Thus, precise and comprehensive annotation across diverse text samples is crucial for effective supervised fine-tuning of these models. 
\item While traditionally the Linguists or Quality Managers are skilled at language translation tasks, to label or annotate PII for LLM SFT, the precise guidelines include understanding of the formats of these PIIs in the specific geographies, e.g., formats of \texttt{PASSPORT NUMBER} or \texttt{LICENSE PLATE} in Poland vs. Vietnam as well as locale-specific nuances such as \texttt{TAX ID NUMBER} or \texttt{TIN} in Polish, Portuguese or Chinese is equivalent to \texttt{PAN ID} in Hindi. 
\item As PII annotation for LLM SFT is emerging, task complexity,  cognitive load, and repetitiveness of these tasks add to annotator fatigue and lead to quality challenges as well as much rework.

\end{enumerate}

\subsection{Key Contributions}
\label{sec:introduction.contributions}

The key contributions of this work is as follows:

\begin{enumerate}
\item Our work targets multilingual annotation of 336 PII types, including passport numbers, bank accounts, health IDs, and driver's licenses specific to each locale (Section~\ref{sec:background.pii_types}).
\item This analytics-driven, phased approach resulted in iterative improvements in Recall and FPR, significantly enhancing data quality (Section~\ref{sec:background.agreement} and \ref{sec:background.metrics}).
\item PII labeling was executed in pilot, training, and production phases, with iterative feedback used to improve annotator performance and data quality (Section~\ref{sec:phased_execution_PII_annotation}).
\item We address both human and analytical aspects of annotating over 40,000 text samples across 13 underrepresented locales. To ensure realistic yet diverse PIIs for LLM training, we employed a semi-automatic process in which synthetic PIIs\footnote{Generated using libraries such as Faker and Mimesis, extended with locale-specific rules.} were incorporated into HiTL-generated prompts. (Section~\ref{sec:annotation_workflow}).
\item Key contributions include computation of inter-annotator agreement, Recall, False Positive Rates (FPR), and root-cause analysis (RCA) of annotation errors and ambiguities (Section~\ref{sec:results.errors}).

\end{enumerate}

\section{Related Work}
\label{sec:relatedwork}
\paragraph{PII Annotation in Multilingual and Low-Resource Settings}
Identifying PII in languages beyond English poses significant challenges due to data scarcity and linguistic diversity. Recent research has explored cross-lingual transfer learning as a solution, such as \cite{amin2022crosslingual}, who achieved substantial performance gains using multilingual BERT on Spanish–Catalan clinical notes. Similarly, \cite{byamugisha2024pii} applied deep neural networks with attention mechanisms to Luganda, demonstrating effective techniques for extreme low-resource settings. Initiatives like \cite{bigcode2023} and \cite{ai4privacy2023} have provided new multilingual datasets, yet comprehensive coverage of low-resource languages remains limited. 

\paragraph{Named Entity Recognition for PII Extraction}
PII annotation is frequently framed as a specialized Named Entity Recognition (NER) task \cite{li2020survey}. However, multilingual contexts pose unique difficulties due to diverse scripts, tokenization methods, and contextual ambiguities \cite{limisiewicz2023tokenization}. Resources such as \textit{WikiANN} (PAN-X) have advanced multilingual NER, leveraging language-agnostic transformer models like XLM-R and multilingual BERT \cite{pan2017cross}. Nonetheless, significant gaps remain, especially in recognizing context-dependent or morphologically complex PII \cite{dernoncourt2017identification}.

\paragraph{Human-in-the-Loop Annotation Frameworks}
Human-in-the-loop (HITL) annotation approaches significantly enhance data quality in sensitive tasks like PII annotation. \cite{liu2023web} illustrate the effectiveness of interactive learning loops, integrating machine predictions with expert verification, leading to substantial improvements in annotation speed and accuracy. HITL frameworks effectively manage the complexity and ambiguity inherent in multilingual PII annotation tasks \cite{settles2011closing}, yet require careful design to avoid introducing annotator biases \cite{settles2011closing}. Studies such as MasakhaNER \cite{adelani2021masakhaner} demonstrate that involving expert annotators with iterative consensus approaches can yield high inter-annotator agreement, achieving Fleiss’ kappa scores near 1.0.

\paragraph{Annotator Agreement and Data Quality for LLM Fine-Tuning}
Consistent annotations are crucial for reliable LLM fine-tuning, particularly when handling sensitive labels such as PII \cite{klie2024analyzing}. Inter-annotator agreement (IAA) is widely used to assess annotation reliability and clarity of guidelines. For example, datasets like CRAPII employ multiple annotators and expert adjudication, highlighting the importance of iterative guideline refinement and annotator training \cite{holmes2024cleaned}. Recent research emphasizes controlling FPR and maximizing recall to ensure robust annotation quality for sensitive contexts such as PII annotation \cite{keymakr}. Rigorous quality control processes are emphasized in the literature as critical for maintaining annotation accuracy and reliability for downstream LLM tasks \cite{klie2024analyzing, keymakr}.

\section{Background}
\label{sec:background}
In this section, we present the execution of the PII annotation, carried out in three independent phases, and outline the annotation workflow alongside formal definition of PII and PII types.

\begin{figure*}[t]
  \centering
  \includegraphics[width=0.48\linewidth]{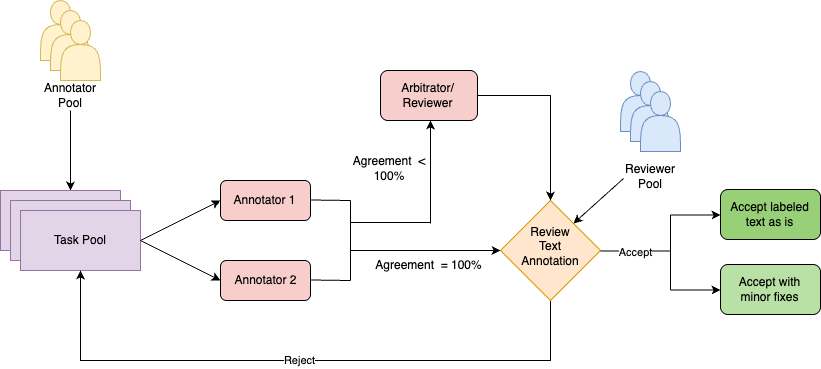}
  \hfill
  \includegraphics[width=0.48\linewidth]{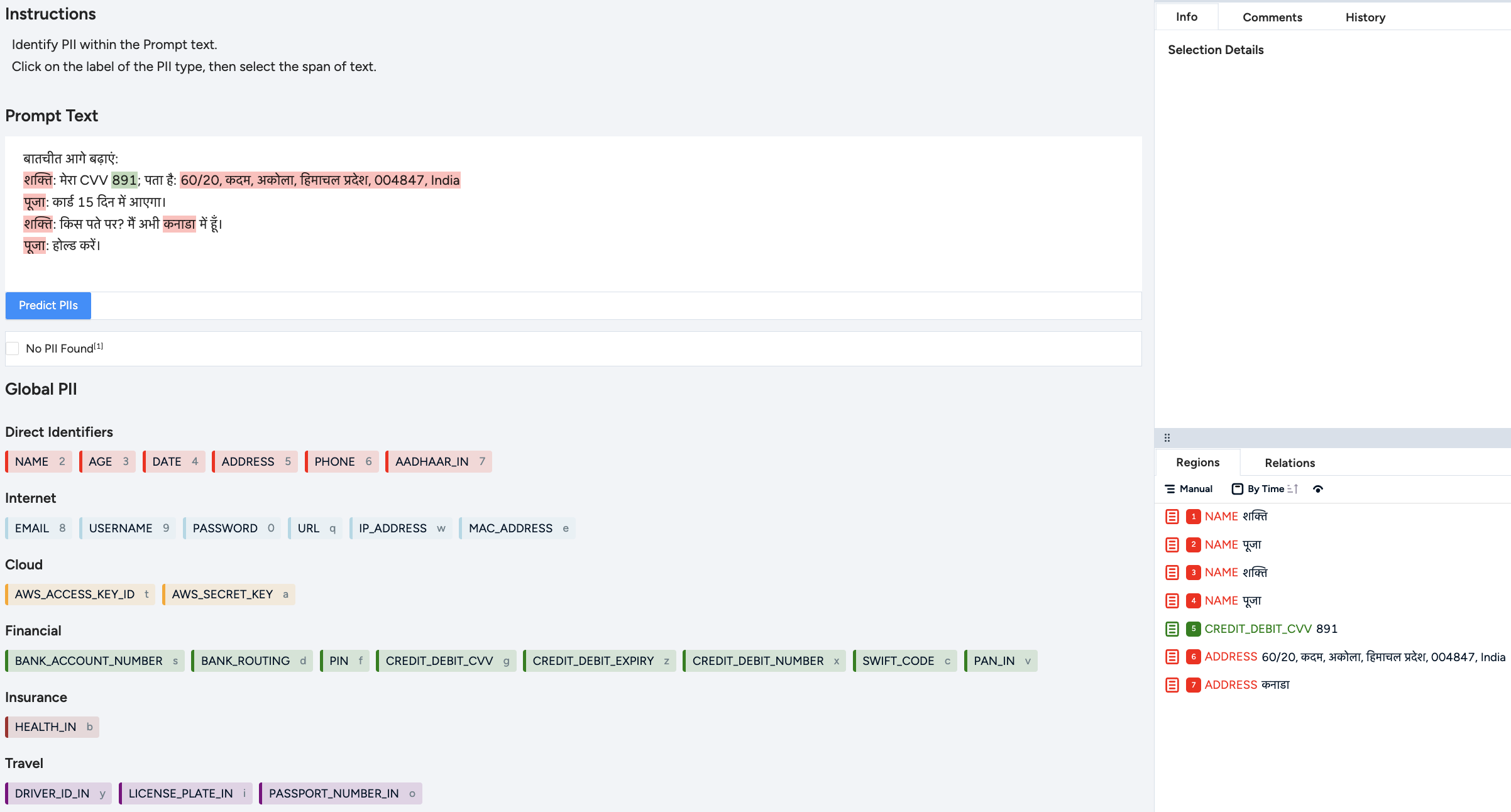}
  \caption{
    Annotator workflow. (Left)  
    User Interface of labeled prompt.(Right) 
  }
  \label{fig:ann_workflow}
\end{figure*}

\subsection{PII definition and locale-specific variations }
\label{sec:background.pii_types}

Personally Identifiable Information (PII) encompasses any data that can be used to uniquely identify an individual. Although there are universal categories, such as \texttt{NAME}, \texttt{AGE}, \texttt{AWS ACCESS KEY ID}, \texttt{EMAIL} that has precise format while \texttt{SSN}, \texttt{NATIONAL ID} , \texttt{HEALTH ID}, \texttt{DRIVER ID} PII format often vary across geographic, cultural, and regulatory contexts. These locale-specific variations pose a significant challenge in building scalable, accurate, and geographically sensitive PII annotation systems. 

Table \ref{tab:pii_types} mentions the different types of PII we used in the study. PII types that are marked with \textbf{*} indicates that these PII types are different for each locale. For example, \texttt{AADHAR ID}, \texttt{PAN ID} is used in Hindi locale while \texttt{NATIONAL ID}, \texttt{SSN} is used in Dutch, Finnish and \texttt{NATIONAL ID}, \texttt{SSN}, \texttt{TIN} is used in Polish, Portuguese, Chinese.

\begin{table}[b]
\centering
\resizebox{\columnwidth}{!}{%
\begin{tabular}{|l|l|l|}
\hline
\multicolumn{3}{|c|}{\textbf{PII Types}} \\
\hline
ADDRESS & AGE & AWS ACCESS KEY ID \\
AWS SECRET KEY & BANK ACCOUNT NUMBER$^{*}$ & BANK ROUTING$^{*}$ \\
CREDIT DEBIT CVV & CREDIT DEBIT EXPIRY & CREDIT DEBIT NUMBER \\
DATE & DRIVER ID$^{*}$ & EMAIL \\
HEALTH ID$^{*}$ & IP ADDRESS & LICENSE PLATE$^{*}$ \\
MAC ADDRESS & NAME & NATIONAL ID$^{*}$ \\
PASSPORT NUMBER$^{*}$ & PASSWORD & PHONE \\
PIN & SSN$^{*}$ & SWIFT CODE \\
TIN$^{*}$ & URL & USERNAME \\
\hline
\end{tabular}%
}
\vspace{2mm}
\caption{List of PII types used in the study.}
\label{tab:pii_types}
\end{table}

\subsection{Phased Execution of PII Annotation}
\label{sec:phased_execution_PII_annotation}
The human-in-the-loop (HITL) annotation of this work was deployed across 13 underrepresented languages, targeting around 336 PII types. Table \ref{tab:localevolumes} summarizes the volume of annotated tasks in the three-phase pipeline.

The challenges associated with multimodal PII annotation were systematically organized into three sequential phases: \textit{Pilot}, \textit{Training}, and \textit{Production}.

\textbf{Pilot Phase:} The first phase functioned as a diagnostic stage to identify early annotation issues. The initial small-volume data set served to uncover locale-specific challenges, PII type ambiguities, and uncertainties in the annotation workflow. Annotators encountered difficulties distinguishing between similar PII types and interpreting culturally specific identifiers. Insights gained during this phase directly informed updates to the annotation guidelines and the development of targeted training materials.

\textbf{Training Phase:} The second phase was designed to develop annotators' proficiency and ensure consistency across the annotation team. Building on the insights from the pilot phase, the training materials were enhanced to support accurate PII labeling across a significantly larger volume of tasks. Any recurring issues, such as annotation disagreements or mislabeling, were addressed by the internal Quality Team through iterative feedback loops, targeted performance reviews, and further refinements to the guidelines, ensuring alignment and calibration across all Annotators.

\textbf{Production Phase:} The third phase involved large-scale quality-assured annotation work conducted under standardized guidelines. At this final stage, the annotation process was scaled to meet the full target volumes. By this point, protocols were well established, and Annotators had gained proficiency in managing complex, locale-specific prompts containing numerous PIIs. Throughout production, consistently high annotation quality was maintained thanks to ongoing monitoring, arbitration, and feedback mechanisms carried out by the internal Quality Team.

The PII annotation work in three independent phases allowed gradual improvements of the learning material and continuous enhancements of the Annotators work, where not only linguistic, but also cultural nuances were correctly taken into account. As a result, the three-phase pipeline developed for the PII annotation work enabled more effective resource allocation, gradual increase in the annotation quality, and reliable results with a larger dataset. Moreover, the Quality Team could develop in-house solutions for challenging PII annotation cases flagged by Annotators and perform necessary Root Cause Analysis with reliable solutions.

\subsection{PII Annotation Process}
\label{sec:annotation_workflow}
The annotation process of PIIs depended on the \textit{Annotators}, \textit{Platform}, \textit{Process}, and \textit{\textit{Ground Truth} Annotation}.

\begin{figure*}[t]
  \centering
  \includegraphics[width=0.48\linewidth]{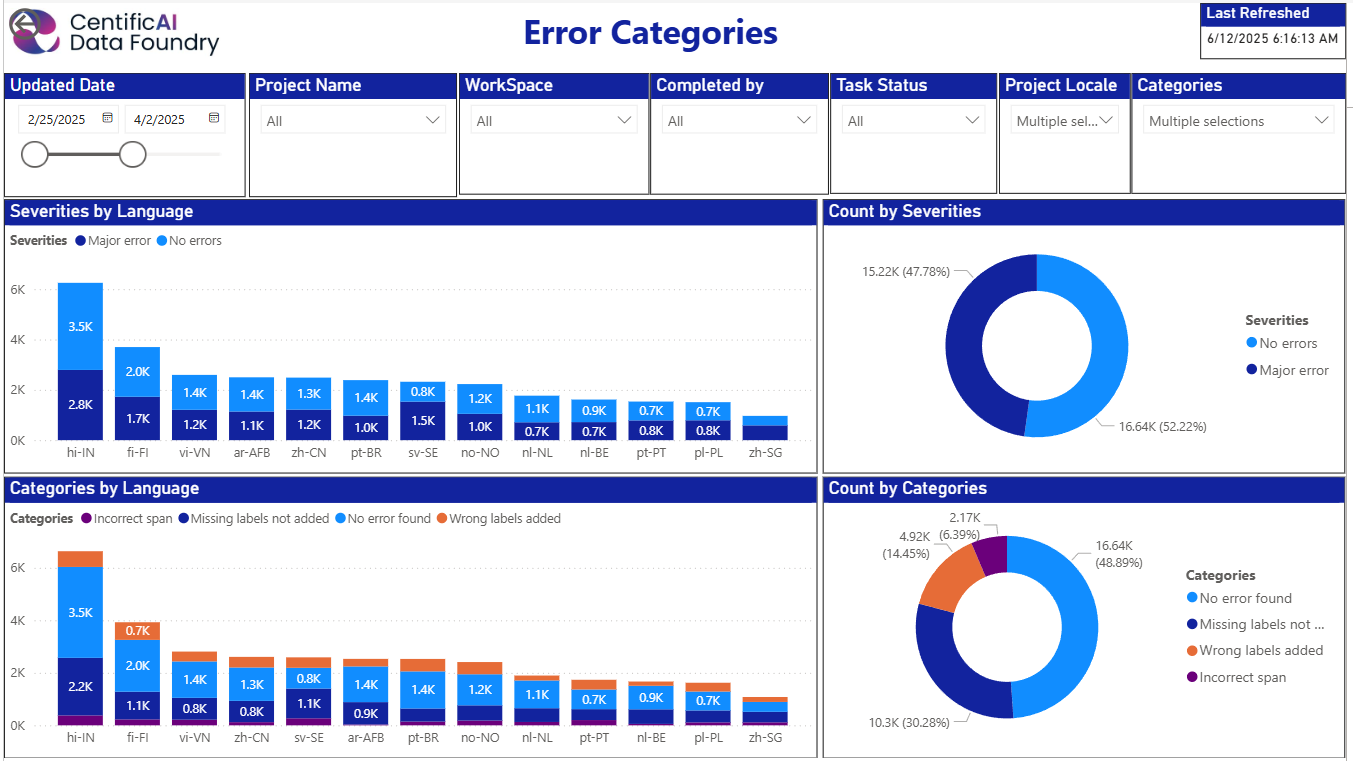}
  \hfill
  \includegraphics[width=0.48\linewidth]{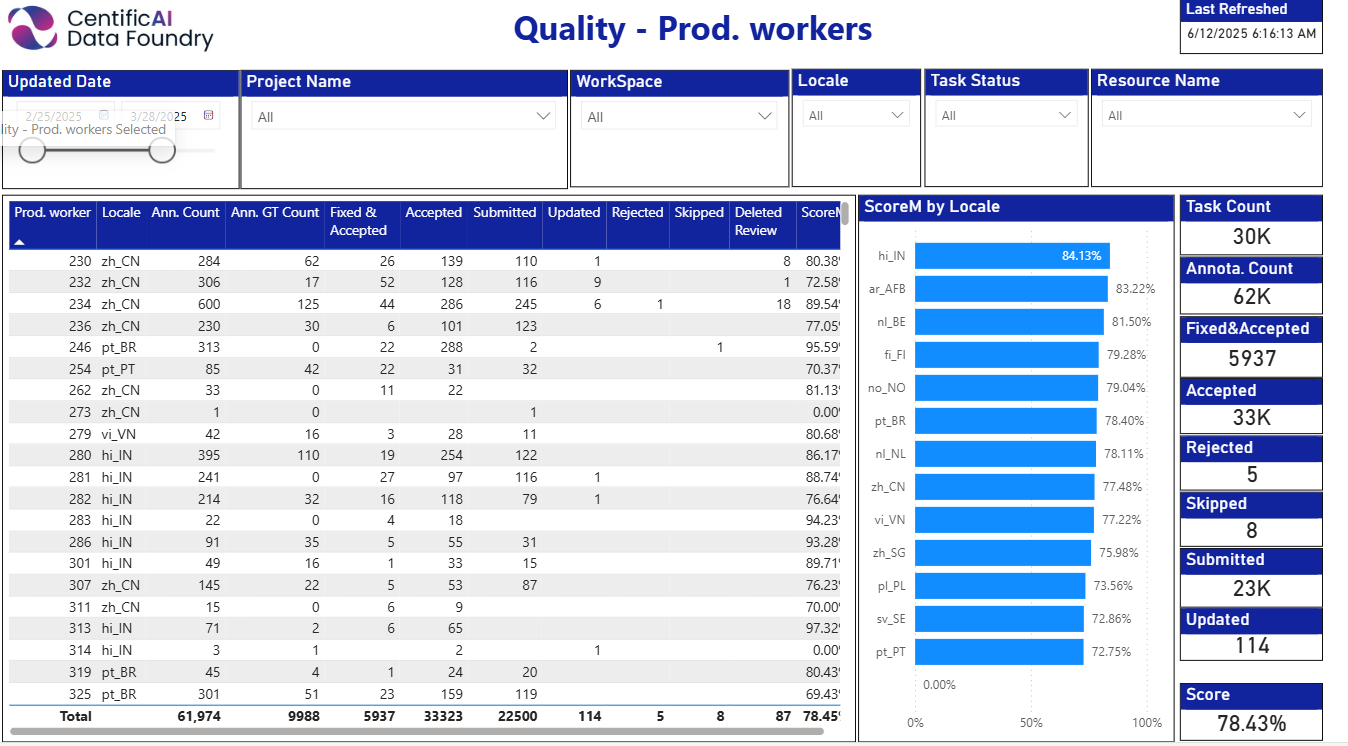}
  \caption{
   Descriptive statistics of \textit{error categories} in PowerBI identified in the QA review per locale. (Left)  
    Quality score of Annotators.(Right) 
  }
  \label{fig:PowerBI}
\end{figure*}

\textbf{Annotators:} The human resources involved in each of the three phases of the annotation work were globally recruited and comprised a group of specialized experts with diverse academic and non-academic backgrounds. Although many Annotators possessed linguistic or large language model (LLM) expertise, PII data curation demanded additional domain-specific training in pre-defined domains (e.g., legal, technical, financial) and a precise adherence to metadata instructions. At the recruitment stage, Annotators underwent a qualification assessment designed to evaluate their understanding of project-specific tasks. Only candidates who achieved a quality score of 85\% or higher were permitted to enter the production work for PII annotation. Furthermore, the annotation work was subject to continuous quality monitoring by a dedicated Quality Assurance (QA) team and subsequently by dedicated Quality Managers (QMs). Based on findings from the quality assessments, Annotators received targeted re-training focused on common error patterns and participated in live FAQ sessions to support their ongoing development and alignment with evolving project standards.

\textbf{Platform:} 
The annotation tasks performed by Annotators were conducted using the DataFoundry platform \cite{HumanSignal:labelstudio}; Fig.~\ref{fig:ann_workflow}, \textit{Right}). In recent years, the emergence of advanced data annotation platforms equipped with features such as SSO access, reusable templates for multimodal data, custom and standard agreement metrics, project configuration tools, progress tracking dashboards, and quality assurance modules has largely replaced the traditional reliance on Excel spreadsheets for annotation workflows. These tailored annotation solutions enable a high degree of customization and leads to more efficient, less human error-prone, and higher-quality annotation output. Features such as inter-rater agreement calculation, dynamic interface configurations, and embedded QA processes contribute to improved governance, scalability, and transparency in managing annotation projects \cite{Centific:AIDF}.

\textbf{Process:} 
Figure~\ref{fig:ann_workflow} illustrates the step-by-step annotation workflow, where each task is assigned to at least two Annotators in the production and then  enabled for the calculation of inter-rater agreement (IRA; Section~\ref{sec:background.agreement}). Based on the percentage agreement, tasks with two submissions per task were either accepted (if IRA fell between 85\% and 100\%, depending on locale) or reassigned for arbitration by an expert Reviewer (QAer). During the QA process, an experienced contributor assessed the available submissions, set the best as the \textit{Ground Truth (GT)}, and determined whether the chosen submission was error-free or required minor adjustments. To facilitate this, the DataFoundry platform \cite{Centific:AIDF} included a dedicated \textit{QA Rubric} section visible only to Reviewers. This section contained a drop-down list of potential \textit{error categories} that the QAer could identify in the GT submissions:

\begin{itemize}
    \item \textit{Missing labels}, when a PII was not spotted in the Annotator's prompt and had to be added by a QAer.
    \item \textit{Wrong labels added}, when the PII label category was not correctly recognized in the production and had to be corrected by a QAer.
    \item \textit{Incorrect span}, when the on- and off-set of the labeled PII span was not correctly set and required QAers' corrections.
\end{itemize}

With the help of error categories, the number of accepted (including corrected and accepted) vs. rejected hits was monitored in PowerBI dashboards (Tab.~\ref{fig:PowerBI}, \textit{Left}). The QA reviews allowed a calculation of Annotators' \textit{Quality Score} (Tab.~\ref{fig:PowerBI}, \textit{Right}) in production and served as a baseline for accuracy and efficiency in the PII labeling and the Annotator's internal assessment. The PowerBI observations allowed the Quality Team to manage the resource pool (e.g., retrain, remove, provide more tasks) and maintain a high level of quality.

\begin{figure*}[t]
  \centering
  \includegraphics[width=0.40\linewidth, height=0.25\textheight]{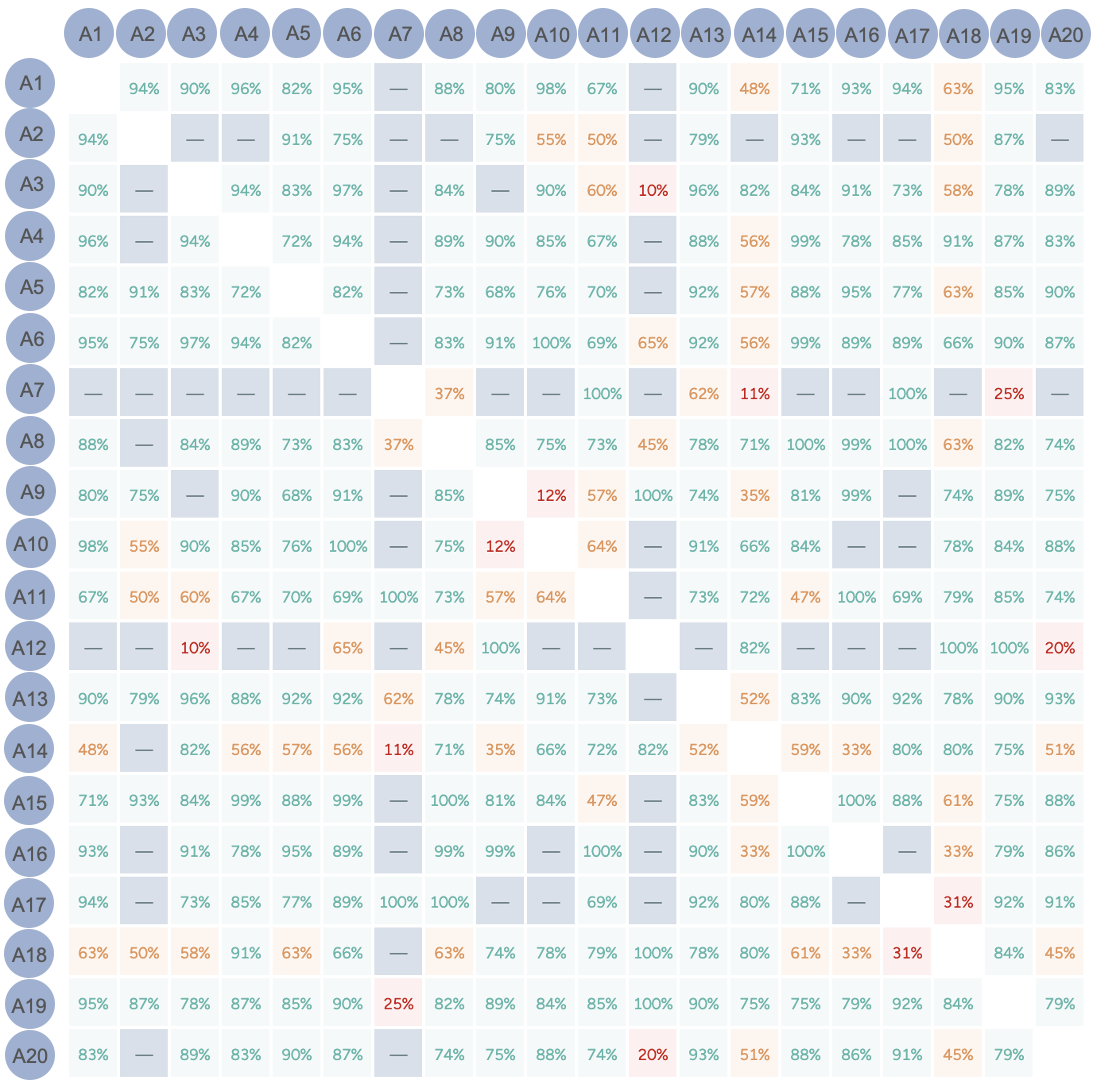}
  \hspace{0.03\linewidth}
  \includegraphics[width=0.40\linewidth, height=0.25\textheight]{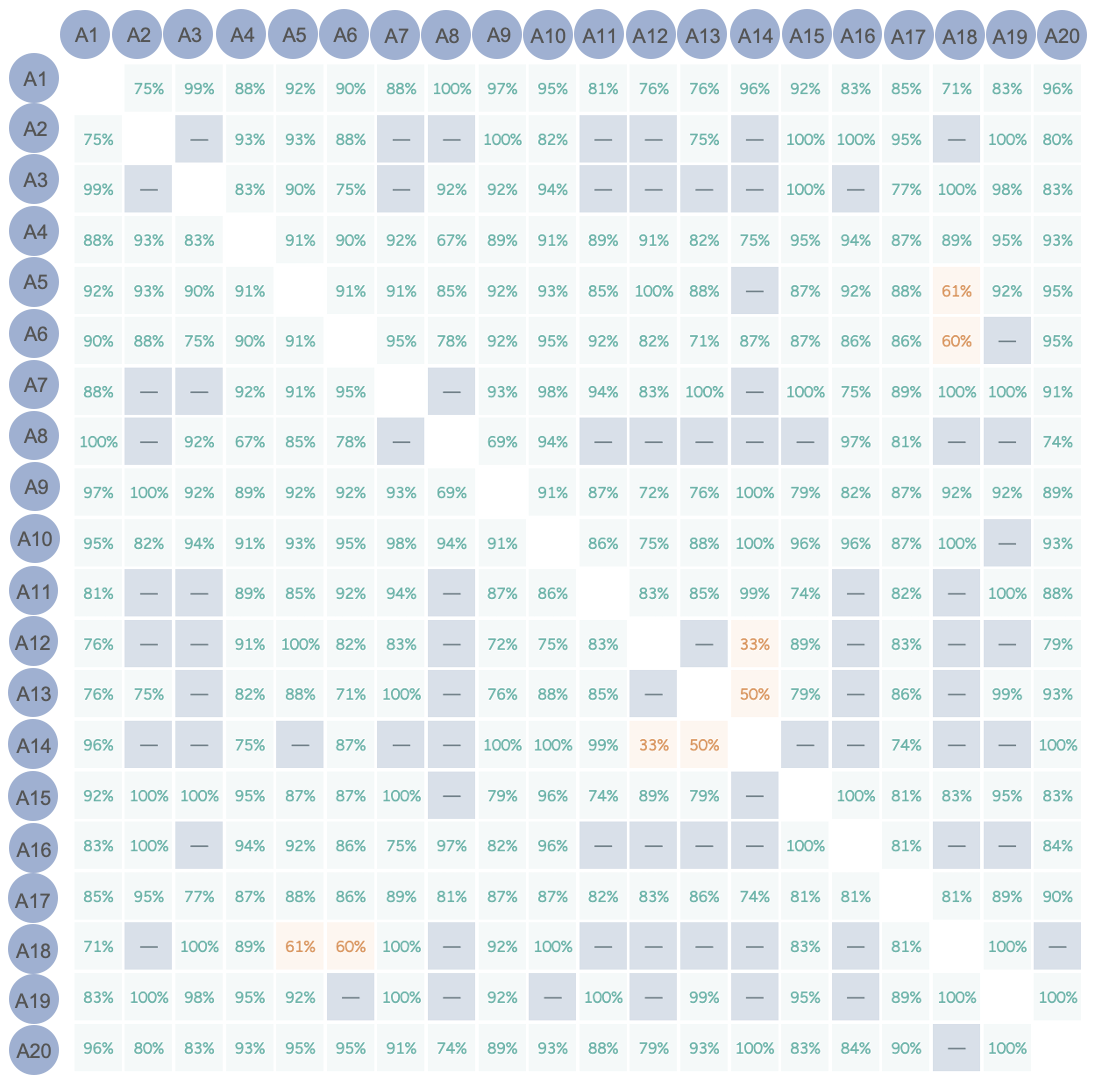}
  \caption{Annotation Matrix for Pilot Phase for pl-PL locale. (Left);
  Production Phase for pl-PL locale. (Right) }
  \label{fig:agreement}
\end{figure*}

\textbf{\textit{Ground Truth} Annotations:} Tasks with two independent Annotators' submissions yielding an inter-rater agreement (IRA) below 85\% were subject to the arbitration process. The involvement of expert QAers varied depending on the phase. Specifically, during the Pilot phase, 100\% of the flagged cases were independently reviewed by QAers regardless of the percentage agreement, since the data subset was relatively small compared to later phases. Additionally, QAers addressed in the Training phase 50\% to 80\% of the disagreements depending on the specific locale. In the Production phase, a subset of 10–15\% of the data underwent QA review due to the time-consuming and costly nature of the process, as well as the large volume of tasks produced.

The disagreements and thus a low value of the inter-rater agreement metrics were a result of two submissions failing at marking correctly the following aspects:

\begin{itemize}
    \item \textbf{PII SPAN:} The \texttt{PII\_START}, \texttt{PII\_END} of a PII information were differently labeled showing clear \texttt{PII\_SPAN} disagreements.
    \item \textbf{PII TEXT:} The labeled PII information was inaccurate (e.g., 25-Aug, 546, username46754).
    \item \textbf{PII TYPE:} The type of PII labeled was incorrectly classified (e.g., \texttt{SSN}, \texttt{PIN}, \texttt{USERNAME}).
\end{itemize}

Additionally, during the production work some Annotators answered that prompts have \textbf{No PII Found}, where there was a clear instance of an PII information. These discrepancies led also to low IRA results. Figure~\ref{fig:ann_workflow}(\textit{Right}) from the hi-IN data set presents a case where an Annotator correctly identified the PIIs: \texttt{NAME}, \texttt{CREDIT DEBIT CVV}, and \texttt{ADDRESS}.

\section{Methodological Approach}
\label{sec:approach}

This section outlines the annotator agreement criteria, accuracy metrics, and data distribution used in the PII annotation process.

\subsection{Agreement Definition}
\label{sec:background.agreement}

\textbf{Task agreement}, often referred to as annotation consensus or labeling consensus, measures the level of agreement between multiple annotators' assigned to the same task \cite{labelstudio2025agreement}. It serves as an important indicator of annotation quality and reliability. This agreement can be evaluated at two levels:
\begin{itemize}
    \item \textbf{Per-task agreement}, which reflects how consistently different annotators' label a single task.
    \item \textbf{Inter-annotator agreement} tells how well the annotations from specific annotators' agree with each other in general, or for specific tasks.
\end{itemize}

The state-of-the-art agreement metrics for PII annotation is defined in Table \ref{tab:agreement_metrics}.

\begin{table}[b]
\centering
\resizebox{\columnwidth}{!}{%
\begin{tabular}{|p{2.8cm}|p{1.2cm}|p{2.8cm}|p{2.8cm}|p{3.2cm}|}
\hline
\textbf{Agreement Metric} & \textbf{Tag} & \textbf{Labeling Type} & \textbf{Description} \\
\hline
Intersection over 1D Regions & Labels & Semantic Segmentation, NER & Checks if spans overlap between Annotators. Used for PII Span. \\

\hline
Matched Spans by (intersection over union) IOU & Labels & Semantic Segmentation, NER & Compares spans by IOU threshold for agreement. Used for PII Span. \\

\hline
Common Label Matches & Taxonomy & Classification, NER & Matches labels or PII types across annotations.  Used for PII Type, PII Text. \\

\hline
\end{tabular}%
}
\vspace{2mm}
\caption{Agreement Metrics for PII annotation \cite{labelstudio2025agreement}.}
\label{tab:agreement_metrics}
\end{table}

We use the mean average of all inter-annotation agreement scores for each annotation pair as the final task agreement score and aggregated the results in an \textit{Annotator Agreement Matrix} in Figure~\ref{fig:agreement}, which shows how agreement scores across all annotations for a task are combined to form a single inter-annotator agreement score in a tabular form. Each cell in the pl-PL matrix obtained in the Production Phase reflects how often a pair of annotators agreed on a particular annotation. The final \% IRA score across two annotators includes agreement in \texttt{PII\_TYPE, PII\_SPAN, and PII\_TEXT}.

\subsection{Accuracy Metrics}
\label{sec:background.metrics}


The two primary metrics calculated are \textit{FPR} and \textit{Recall}. The choice of using these metrics is motivated by the practical implications of PII misclassifications in real-world applications. Recall is critical because it reflects the model’s ability to identify all actual PII instances missing a PII entity (false negative) that can lead to serious privacy breaches. Conversely, FPR captures the rate at which non-PII content is incorrectly labeled as PII, which is important for ensuring the utility and readability of the redacted or labeled output. While metrics like Precision and F1-score are also informative, our use case prioritizes minimizing undetected PII (i.e., high Recall) and reducing unnecessary redactions (i.e., low FPR), which are more aligned with user trust and compliance goals in privacy-sensitive environments.

The calculation of the accuracy metrics was based on the \texttt{PII\_TYPE} comparison of the QAer's \textit{GroundTruth} labels with the Annotator labels and categorized depending on the output into three categories: 

\begin{itemize}
    \item \textbf{agreement}, when the labels are matching,
    \item \textbf{disagreement}, when the labels are not matching,
    \item \textbf{N/A}, when the task is not reviewed by a QAer.
\end{itemize}

This taxonomy allowed us to compute FPR and Recall rates accurately and consistently across data sets from the three phases.

\textbf{FPR} measures the proportion of non-PII types (negative cases) that are incorrectly labeled as \texttt{PII\_Type} (positive cases), relative to the total number of actual negative cases. The comparison of PIIs between an Annotator and a QA Reviewer is considered a False Positive, when a \texttt{PII\_Type} is annotated by one but not the other, or when any one of them assign different \texttt{PII\_Type}.

\begin{equation}
\makebox[\columnwidth][l]{%
\resizebox{\columnwidth}{!}{$
\text{FPR} = \frac{\text{False Positives}}{\text{False Positives} + \text{True Negatives}} = \frac{\text{False Positives}}{\text{All Actual Negatives}}
$}
}
\end{equation}

\textbf{Recall} defined as the proportion of true \texttt{PII\_Type}(s) correctly identified out of all actual \texttt{PII\_Type}(s). A case is considered a True Positive when the PII annotated by both the Annotator and the QA Reviewer exactly matches the \texttt{PII\_Type} in the ground truth, including both type and count in the correct order.
\begin{equation}
\makebox[\columnwidth][l]{%
\resizebox{\columnwidth}{!}{$
\text{Recall} = \frac{\text{True Positives}}{\text{True Positives} + \text{False Negatives}} = \frac{\text{Correctly Retrieved Positives}}{\text{All Actual Positives}}
$}
}
\end{equation}

FPR and Recall metrics are computed at two levels: 

\textbf{Coarse-grained}: Metrics are computed at the row level, where a row is considered correct if at least one \texttt{PII\_TYPE} is correctly identified, regardless of the total number or order of PII types in that row.

\textbf{Fine-grained}: Metrics are computed at a more detailed level, where a row is considered correct only if all \texttt{PII\_TYPE}s are correctly identified in the exact order and with the same count as in the ground truth.

The metrics computation followed a standardized pipeline entailing data normalization (i.e., handling NAs, standardizing labels), fine-grained analysis and coarse-grained analysis. This process ensured rigorous and reproducible evaluation of annotation quality. For the purposes of this study, only fine-grained metrics are considered.

\begin{figure*}[t]
  \centering
  \includegraphics[width=0.48\linewidth]{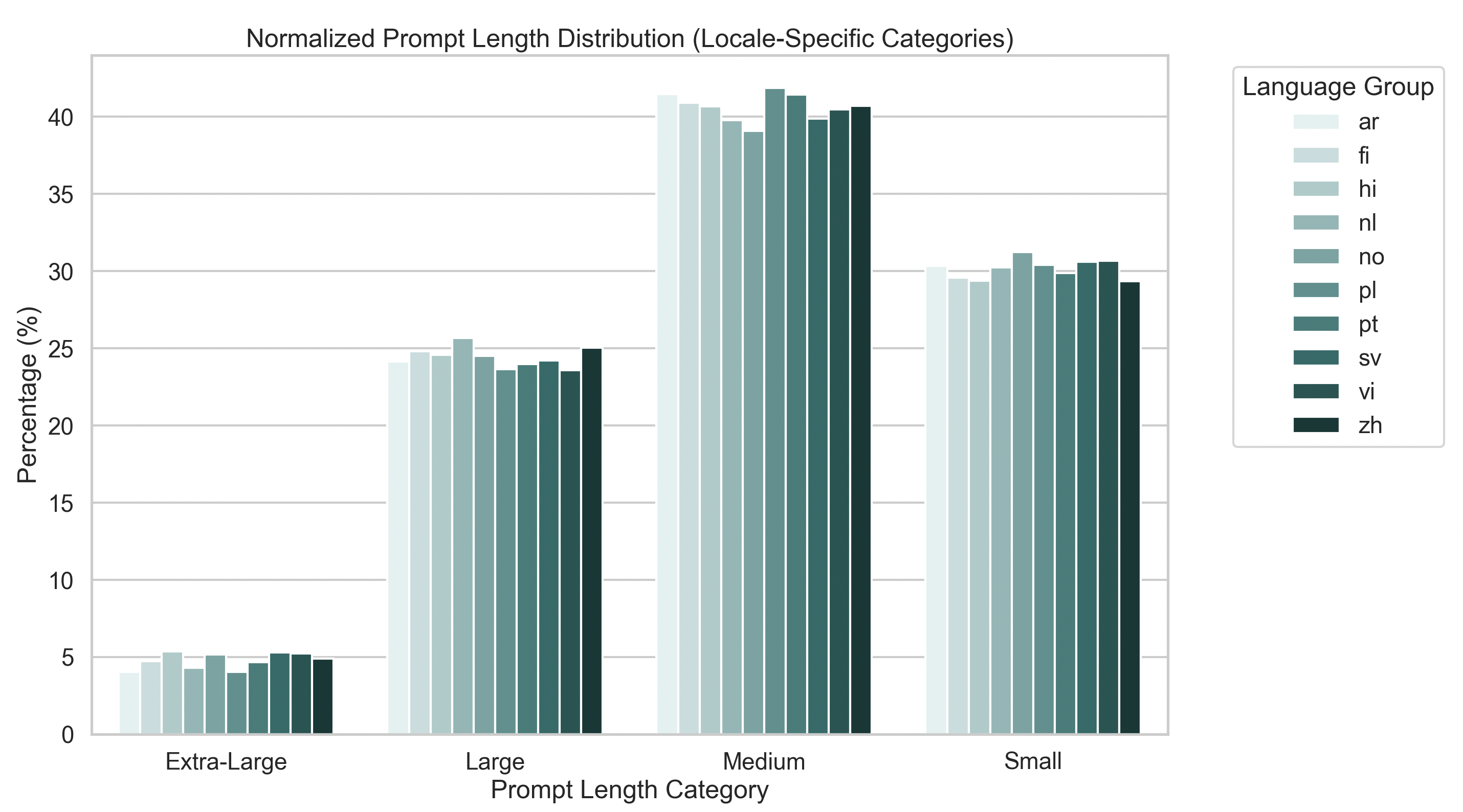}
  \hfill
  \includegraphics[width=0.48\linewidth]{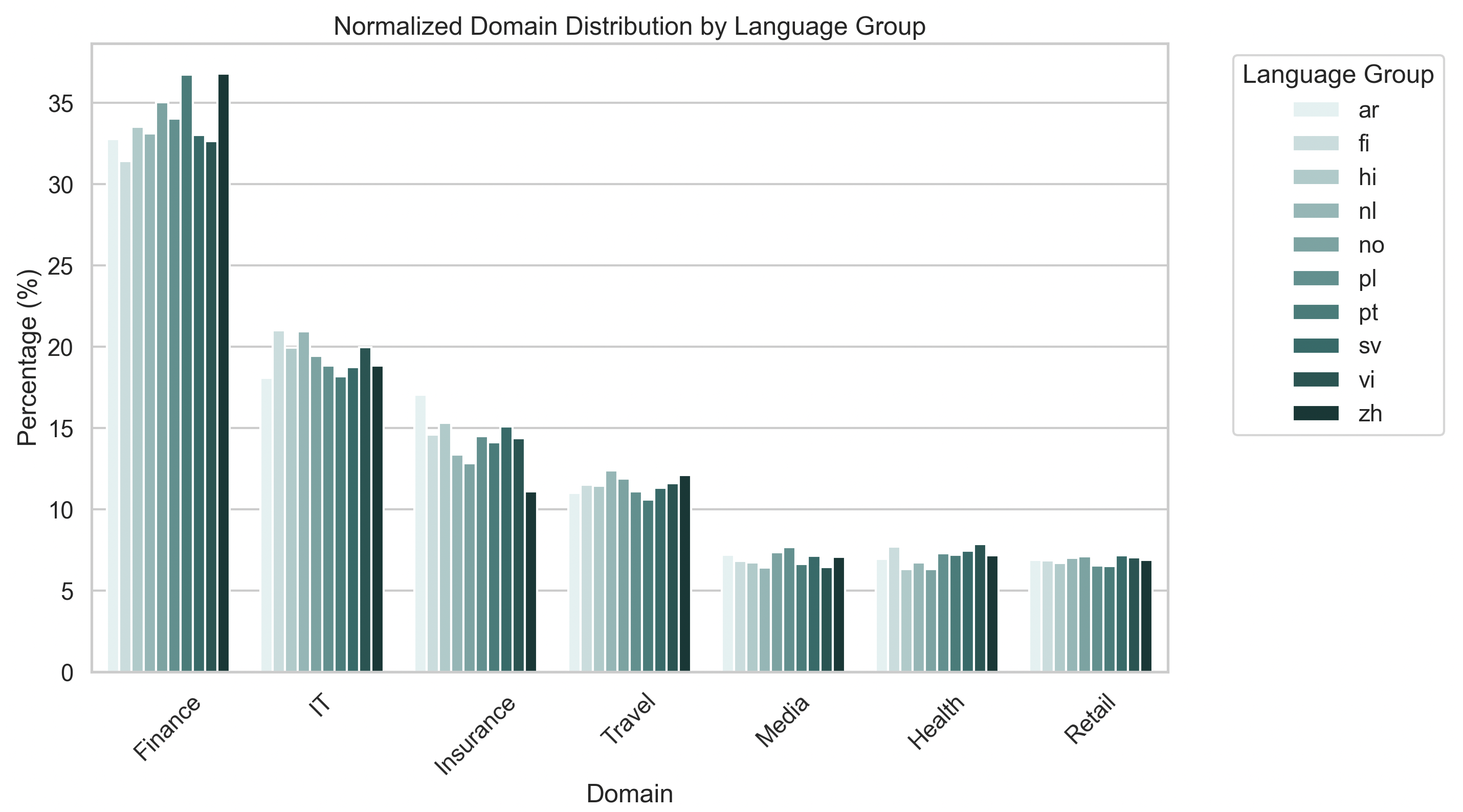}
  \caption{
    Normalized domain distribution across locales. (Left)  
    Normalized prompt length distribution across locales.(Right) 
  }
  \label{fig:domain_prompt}
\end{figure*}

\begin{figure}[t]
  \centering
  \includegraphics[width=\columnwidth]{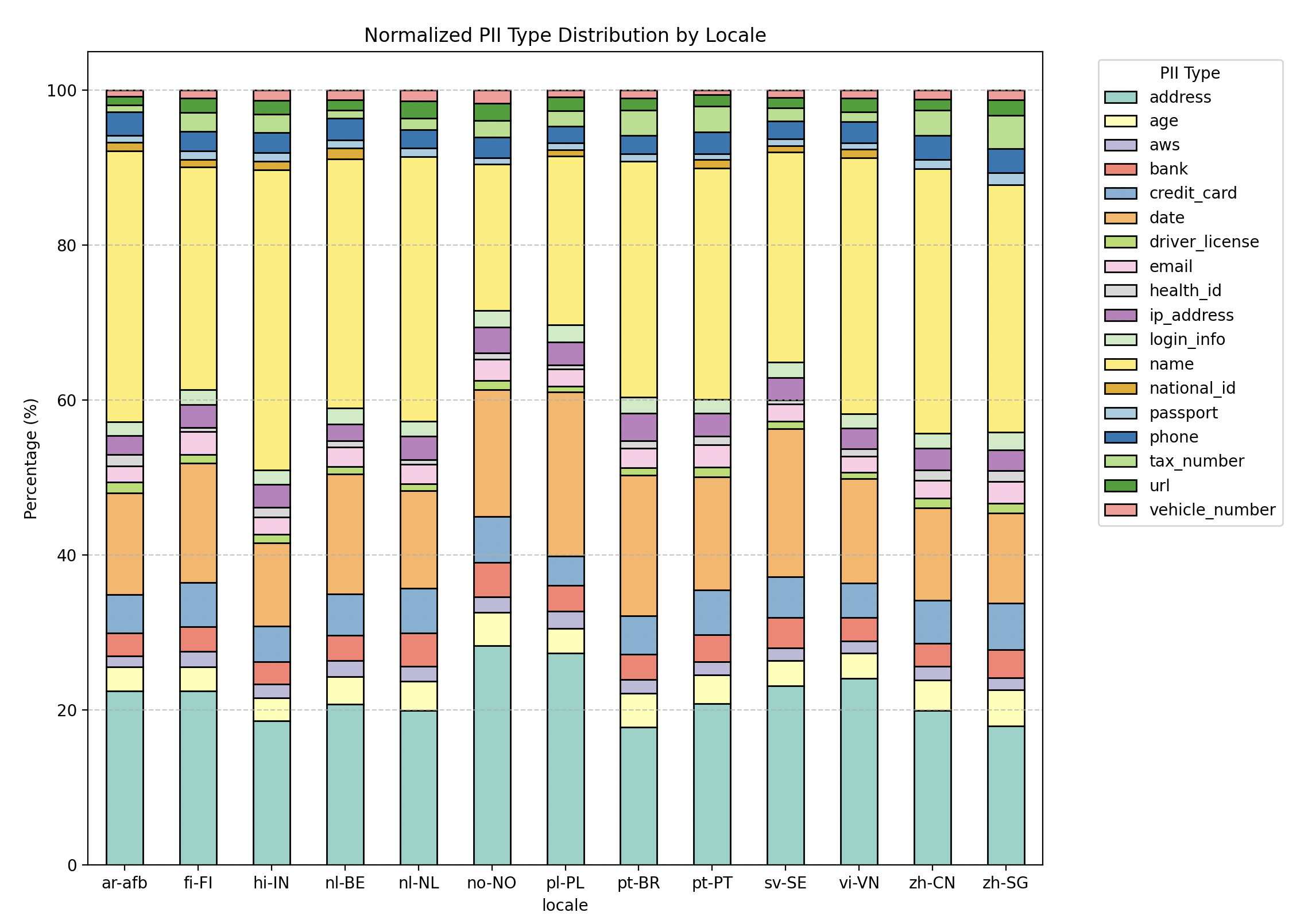}
  \caption{Normalized distribution of PII types across 13 locales.}
  \label{fig:pii_type_distribution}
\end{figure}

\subsection{Dataset Distribution}
\label{sec:data_prep.data_dist}

We present a detailed breakdown of our dataset in multiple dimensions, prompt length, domain representation, and PII-type distribution, to highlight its diversity and coverage.

\textbf{Domain Distribution:} To ensure the reliability and fairness of our multilingual dataset, we observed that the coverage of the domains encompassing categories such as finance, health, IT, insurance, media, retail and travel remains consistent between languages, with no single domain represented disproportionately, as shown in Figure~\ref{fig:domain_prompt}. This uniformity supports the claim that domain-specific content is equitably distributed across different linguistic contexts.

\textbf{Prompt Length Distribution:} For prompt length, we classified inputs into four bins: Small (S), Medium (M), Large (L), and Extra-Large (XL), where size thresholds were determined separately for each locale, based on language-specific variations in sentence structure and verbosity. In general, Small corresponds to fewer than 30 words, Medium to 30 to 200 words, Large to 240 to 1,200 words, and Extra Large to 1,200 to 3,500 words. Despite this language-specific calibration, the relative distribution of prompt lengths across categories remains consistent across languages, reflecting a balanced representation of input complexity between languages as shown Figure~\ref{fig:domain_prompt}.

\textbf{PII Type Distribution:} To simplify visualization and reduce complexity in analysis, semantically similar PII categories were merged, reducing the total from 28 to 18, as detailed in Table \ref{tab:pii_type_mapping}. Figure~\ref{fig:pii_type_distribution} presents the normalized distribution of PII types across various locales. A prominent trend is the universal dominance of the name entity, consistently comprising the largest proportion of detected PII, often exceeding 30\% of the total in each locale. \texttt{Address} and \texttt{Date} PII types also demonstrate widespread prevalence, although their proportions vary by locale. For example, locales such as ar-AFB, fi-FI, vi-VN, no-NO, pl-PL and sv-SE exhibit higher \texttt{Address} frequencies. An equal distribution is observed across the PII types: \texttt{Phone}, \texttt{Login Info}, \texttt{IP Address}, \texttt{Email}, \texttt{Bank}, \texttt{Age}, and \texttt{Credit Card}.

\begin{table}[b]
\centering
\renewcommand{\arraystretch}{1.2}
\resizebox{\columnwidth}{!}{%
\begin{tabular}{|l|l|}
\hline
\textbf{PII Type Category} & \textbf{PII Type Mapping} \\
\hline
CREDIT CARD & CREDIT CARD NUMBER, CVV, PIN \\
\hline
TAX NUMBER & TIN, SSN, PAN \\
\hline
BANK & BANK ACCOUNT NUMBER, SWIFT CODE, INTERNATIONAL BANK ACCOUNT NUMBER \\
\hline
HEALTH ID & HEALTH ID \\
\hline
DRIVER LICENSE & DRIVER ID \\
\hline
PASSPORT & PASSPORT NUMBER \\
\hline
NATIONAL ID & NATIONAL ID, AADHAR ID \\
\hline
VEHICLE NUMBER & LICENSE PLATE\\
\hline
IP ADDRESS & IP ADDRESS, MAC ADDRESS \\
\hline
AWS & AWS ACCESS KEY, AWS ACCESS KEY ID \\
\hline
URL & URL \\
\hline
NAME & NAME \\
\hline
PHONE & PHONE NUMBER \\
\hline
EMAIL & EMAIL ADDRESS \\
\hline
AGE & AGE \\
\hline
DATE & DATE \\
\hline
LOGIN INFO & USERNAME, PASSWORD \\
\hline
\end{tabular}%
}
\vspace{2mm}
\caption{Mapping of similar PII types to a single PII Type Category}
\label{tab:pii_type_mapping}
\end{table}

\begin{figure*}[t]
  \centering
  \begin{minipage}[b]{0.48\textwidth}
    \centering
    \includegraphics[width=\textwidth, height=0.4\textheight, keepaspectratio]{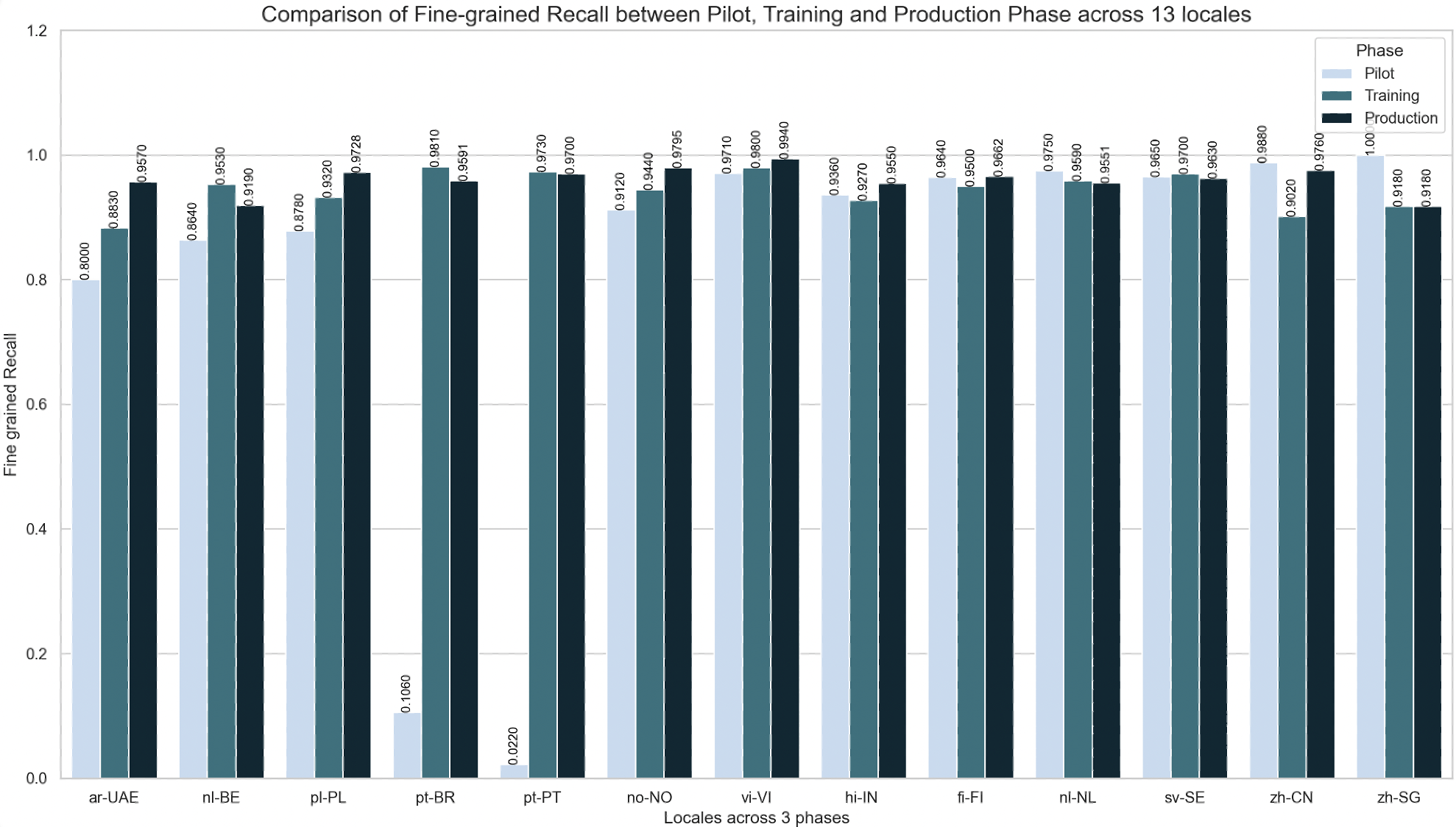}
  \end{minipage}%
  \hfill
  \begin{minipage}[b]{0.48\textwidth}
    \centering
    \includegraphics[width=\textwidth, height=0.4\textheight, keepaspectratio]{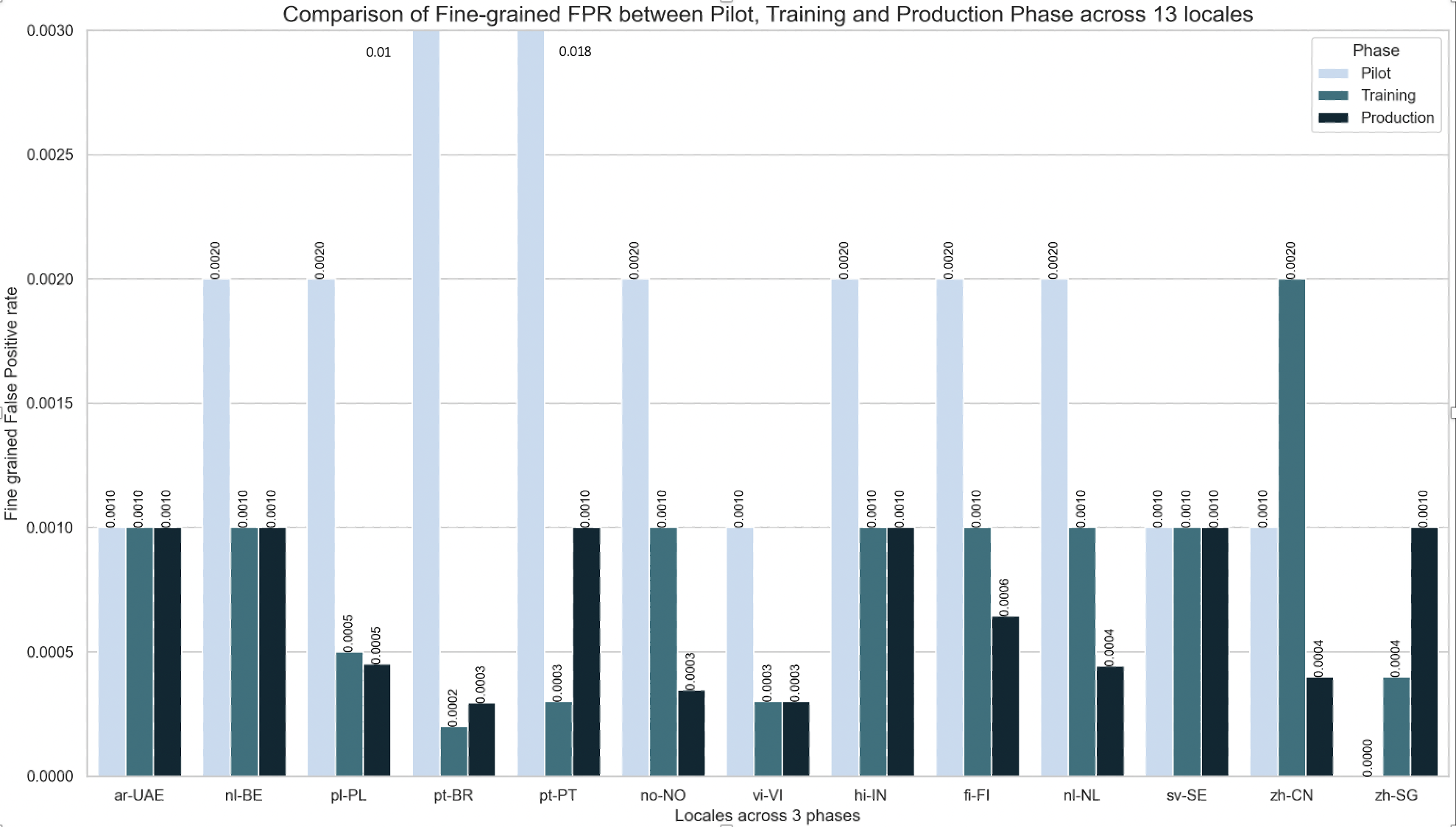}
  \end{minipage}
  \caption{Comparison of (Left) Fine-grained Recall and (Right) Fine-grained FPR across 13 locales and 3 phases.}
\label{fig:recall_fpr}
\end{figure*}

To support fair comparison and visualization, we normalized all distributions, converting raw counts into proportions. This was essential to mitigate the effects of varying sample sizes across locales and to ensure that cross-locale comparisons reflect true distributional patterns rather than volume-based disparities. Additionally, linguistically similar locales (e.g., nl-BE and nl-NL into nl, pt-BR and pt-PT into pt, zh-CN and zh-SG into zh) were grouped to improve representational consistency for domain and prompt length distribution plots. Overall, these steps confirm that our dataset exhibits no language-specific skew in domain, prompt length and PII type reinforcing its suitability for multilingual evaluation and modeling.

\section{Results and Discussion}
\label{sec:results}

\subsection{PII (dis)agreement, FPR, and Recall}
\label{sec:results.matrix}

The IRA scores between two annotators are computed for PII data collected in each phase (see Section~\ref{sec:background.agreement}). A higher percentage of agreement indicates stronger consistency between annotators. By extension, this suggests that the provided guidelines and supporting materials for the PII labeling task were clearer, thereby improving both data quality and reliability. Figure~\ref{fig:agreement} illustrates how agreement scores evolved over time, from the small-scale pilot to the large-scale production phase. Overall, most locales demonstrated visible improvements, suggesting a steady enhancement in PII detection and annotation quality. 

The calculation of FPR and Recall across multiple locales also changed over time. Specifically, FPR decreased while Recall increased, highlighting improvements in both precision and sensitivity of the PII annotation process.

\textbf{FPR Improvement:}  As shown in Figure~\ref{fig:recall_fpr}, a substantial drop in FPR was observed from pilot to production in locales such as pt-BR (0.015 → 0.0003), pt-PT (0.018 → 0.001), and no-NO (0.002 → 0.0003), with similar trends in nl-BE, pl-PL, vi-VI, hi-IN, fi-FI, and nl-NL. ar-AFB and sv-SE maintained a stable FPR, while zh-SG showed a slight increase (0 → 0.001). In zh-CN, FPR rose during the pilot phase (0.001 → 0.002) and dropped again in production (0.0004).

\textbf{Recall Improvement:} As shown in Figure \ref{fig:recall_fpr}, significant recall gains were observed from the pilot to production phase in locales such as ar-AFB (0.80 → 0.957), pt-BR (0.106 → 0.959), and pt-PT (0.022 → 0.970), with similar improvements seen in nl-BE, pl-PL, no-NO, hi-IN, and vi-VI. Locales such as fi-FI and sv-SE showed stable recall, while a slight decrease was observed in nl-NL, zh-CN, and zh-SG.

For the zh-CN and zh-SG locales, the drop in recall may be attributed to the linguistic characteristics of Chinese, such as the absence of word boundaries and character-level ambiguity. These factors make it more difficult for annotators to consistently identify and label PII spans. nl-NL includes compound words and grammatical inflections, which may introduce ambiguity to annotators during labeling. These linguistic characteristics can lead to inconsistent or partial annotations of PII entities, contributing to a drop in recall. 

These results indicate that iterative improvements across phases effectively reduced FPR and improved recall across 10 locales out of 13 locales from pilot to training to production phase.

\subsection{PII Annotation Disagreements}
\label{sec:results.errors}

The strong agreement achieved in the final Production phase, validated across hundreds of prompts in all 13 locales, was the result of a thorough root cause analysis of the disagreements identified in annotator's work. While such issues were not observed in the Pilot dataset, recurring mistakes began to emerge rapidly during the Training phase.

To address uncertainties in PII labeling and resolve confusions arising from similarly appearing entities, annotators received rigorous training supported by carefully curated learning materials that included both positive and negative examples. Regular Root Cause Analyses (RCA) of annotation disagreements were conducted on a weekly basis to identify systematic issues and refine the annotation guidelines. Furthermore, disagreements between aggregated Annotator PII labels and the QAed Ground Truth submission were examined, categorized, and subsequently resolved. Table~\ref{tab:rca} summarizes the specific cases of disagreements.

\begin{table}[b]
\renewcommand{\arraystretch}{1.2}
\label{tab:rca}
\begin{tabular}{|p{2.5cm}|p{5.3cm}|}
\hline
\textbf{Category} & \textbf{Description} \\
\hline
\texttt{PII\_TYPE} & Mismatches in the classification of PII types (e.g., CVV vs. PIN). \\
\hline
\texttt{PII\_SPAN} & Discrepancies in the start/end positions of PII annotations. \\
\hline
\texttt{PII\_TEXT} & Differences in the actual text content of PIIs (e.g., 123-456-7890 vs. 1234567890). \\
\hline
\texttt{NUMBER OF PIIs} & Mismatches in the number of PIIs identified in a text. \\
\hline
\texttt{SAME PII ORDER} & Variations in the sequence of PII annotations. \\
\hline
\end{tabular}
\vspace{2mm}
\caption{Different root cause analysis of disagreement categories for PII annotation}
\end{table}

The first two categories, \texttt{PII\_TYPE} and \texttt{PII\_SPAN}, were the primary focus in the analysis of annotator disagreements, as they most directly reflect annotation precision and consistency. In contrast, information about NUMBER OF PIIs and SAME PII ORDER served as quick indicators of discrepancies and provided immediate insights into the QA reviewers’ thoroughness and vigilance when evaluating annotator work. Beyond this, the full set of disagreement categories enabled detailed profiling of error types and informed targeted updates to both the guidelines and training materials. The RCA outputs included comprehensive error reports and quantitative metrics, which collectively supported continuous process improvement across all phases.

Table~\ref{tab:pii_errors} highlights the most common annotation errors related to PII types observed globally. For example, \texttt{CREDIT DEBIT NUMBER} was often mislabeled with \texttt{BANK ACCOUNT NUMBER}. Similarily, Annotators had difficulties in separating the social security number (\texttt{SSN}) with the national health number (\texttt{HEALTH ID}) or \texttt{NATIONAL ID} as these were often used interchangeably in specific countries, but should be differently categorized depending on the context of the prompt. Similarly, Annotators struggled to correctly label \texttt{AWS SECRET KEY} and separate it from the \texttt{AWS ACCESS KEY ID} as these are not the commonly used PIIs and is familiar only to a small group of people.

\begin{table}[t]
\centering
\resizebox{\columnwidth}{!}{%
\begin{tabular}{|p{11cm}|}
\hline
\textbf{Pilot Phase} \\ \hline
CREDIT DEBIT NUMBER~$\leftrightarrow$~BANK ACCOUNT NUMBER \\
CVV~$\leftrightarrow$~PIN \\
SSN~$\leftrightarrow$~HEALTH ID \\
TIN~$\leftrightarrow$~SSN \\
AWS SECRET KEY~$\leftrightarrow$~AWS ACCESS KEY ID \\
\hline

\textbf{Training Phase} \\ \hline
USERNAME~$\leftrightarrow$~SWIFT CODE \\
SSN~$\leftrightarrow$~PIN \\
HEALTH ID~$\leftrightarrow$~NATIONAL ID \\
SSN~$\leftrightarrow$~HEALTH ID \\
MAC ADDRESS~$\leftrightarrow$~IP ADDRESS \\
CREDIT DEBIT CVV~$\leftrightarrow$~CREDIT DEBIT NUMBER \\
USERNAME~$\leftrightarrow$~EMAIL \\
PHONE NUMBER~$\leftrightarrow$~BANK ROUTING \\
\hline

\textbf{Production Phase} \\ \hline
BANK ACCOUNT NUMBER~$\leftrightarrow$~CREDIT DEBIT NUMBER \\
DATE~$\leftrightarrow$~CREDIT DEBIT EXPIRY \\
\hline
\end{tabular}%
}
\vspace{2mm}
\caption{Common PII Type Annotation Errors Across Phases}
\label{tab:pii_errors}
\end{table}

\section{Conclusion and Future Work}
\label{sec:futurework}

This work presents a scalable multilingual data curation framework for sensitive PII annotation with an emphasis on quality, inter-annotator agreement, and accuracy metrics. Our approach demonstrates substantial improvements in recall and FPR across locales using a phased approach. 

Below we discuss potential future works:
\begin{enumerate}
\item While we use the \texttt{PII\_TYPE} as the accuracy metric, in future, we will explore using \texttt{PII\_SPAN} and \texttt{PII\_TEXT} as well. This will enable multi-faceted evaluation.
\item We will also explore efficiency gains using AI-assisted PII annotation as a future work, comparing against the state-of-the-art NER solutions \cite{li2020survey}.   
\item In future, we plan to evaluate the impact of the \textit{quality} of our curated multilingual dataset on the \textit{accuracy} achieved by the fine-tuned LLMs. 

\end{enumerate}

\bibliographystyle{IEEEtran}   
\bibliography{IEEEabrv, IEEE}

\end{document}